\documentclass[letterpaper, 10 pt, conference]{ieeeconf}  

\IEEEoverridecommandlockouts 
                                                          
\usepackage{booktabs}
\usepackage[flushleft]{threeparttable}
\usepackage{xcolor}
\usepackage[utf8]{inputenc}
\usepackage{float}
\usepackage{caption}
\usepackage{url,xspace}
\usepackage{array}
\usepackage{balance}
\usepackage{hyphenat}

\usepackage{pifont}
\usepackage{tabularx}
\usepackage{subcaption}
\usepackage[most]{tcolorbox}
\usepackage{multirow}
\usepackage{hyperref}

\makeatletter
\@ifundefined{algorithmautorefname}
  {}
  {}
\makeatother

\usepackage{ifthen}
\newboolean{showcomments}
\setboolean{showcomments}{true}
\ifthenelse{\boolean{showcomments}}
{\newcommand{\nb}[2]{%
  \fcolorbox{black}{gray!20}{\bfseries\sffamily\scriptsize#1:}%
  {\sf\small$\blacktriangleright$\textit{#2}$\blacktriangleleft$}%
}}
{\newcommand{\nb}[2]{}}

\newcommand{\head}[1]{\noindent\textbf{#1.}}

\newcounter{fcounter}
\newcommand\finding[1]{\refstepcounter{fcounter}%
\vspace{2pt}\begin{tcolorbox}[boxsep=1pt,left=2pt,right=2pt,top=1pt,bottom=1pt]%
\noindent\emph{\textbf{Finding \arabic{fcounter}}: #1}%
\end{tcolorbox}\vspace{2pt}}

\usepackage{cite}
\usepackage{amsmath,amssymb,amsfonts}
\usepackage{algorithm}
\usepackage{algpseudocode} 
\usepackage{graphicx}
\usepackage{textcomp}
\usepackage{xcolor}

\makeatletter
\newcommand{\linebreakand}{%
  \end{@IEEEauthorhalign}
  \hfill\mbox{}\par
  \mbox{}\hfill\begin{@IEEEauthorhalign}
}
\makeatother

\begin{document}

\title{Automated Evaluation of Multi-Turn Dialogues in In-Car Conversational Assistants}

\author{Vaishnav Negi$^{1,2}$, Lev Sorokin$^{3,2}$, Soroosh Tayebi Arasteh$^{1,4}$, and Andrea Stocco$^{3,5}$%
\thanks{$^{1}$Friedrich-Alexander-Universit\"at Erlangen-N\"urnberg, Erlangen, Germany
        {\tt\small vaishnav.negi@fau.de}}%
\thanks{$^{2}$BMW Group, Germany}%
\thanks{$^{3}$Technical University of Munich, Munich, Germany
        {\tt\small lev.sorokin@tum.de | \tt\small andrea.stocco@tum.de}}%
\thanks{$^{4}$RWTH Aachen University, Aachen, Germany
        {\tt\small soroosh.arasteh@rwth-aachen.de}}%
\thanks{$^{5}$fortiss GmbH, Munich, Germany
        {\tt\small stocco@fortiss.org}}%
}

\maketitle
\thispagestyle{empty}
\pagestyle{empty}


\begin{abstract}
In-car conversational assistants (ICAs) are increasingly integrated into vehicles to support route planning, vehicle control, and information access. Ensuring their reliability is challenging due to multi-turn interactions, the absence of explicit ground truth, and strict safety constraints. Existing evaluation techniques fall short, as they target single-turn settings and fail to capture constraint handling, context retention, and safety-critical behavior across turns.
We propose an automated framework for testing the multi-turn conversational capabilities of ICAs. The system is treated as a black box and evaluated via closed-loop simulation with a strategy-guided user simulator, an adversarial strategy manager, and a two-tier LLM judge assessing turn-level failures and conversation-level quality.
We evaluate the approach on an industrial ICA with six LLM backends and twelve human annotators. The automated judge shows substantial agreement with humans, and strategy guidance uncovers 2.96$\times$ more unique failure types per conversation and more than doubles the number of unique failing conversations compared to unguided simulation.
\end{abstract}


\section{Introduction}\label{sec:introduction}

In-car conversational assistants (ICAs) based on Large Language Models (LLMs) are becoming an integral part of modern vehicles, facilitating voice interaction for navigation, climate control, media, and vehicle information retrieval~\cite{vogel2018emotion,schmidt2019proactive,giebisch25factual,carexpert,2026-Guo-OJ-ITS}. 
However, these systems require higher reliability due to their critical application domain~\cite{2020-Riccio-EMSE,2020-Humbatova-ICSE}. A violated constraint or an overlooked contextual feature during interaction may reduce driver trust and compromise vehicle safety~\cite{vogel2018emotion}. 

Assessing the reliability of such systems automatically, however, poses three interrelated challenges.
Firstly, such dialogues are \textit{multi-turn}, and users may refine their requests and constraints incrementally, such that errors may not be apparent immediately, especially due to long-range dependencies~\cite{kwan2024mteval,guan2025survey}.
Secondly, there is \textit{no unique ground truth} as navigation requests may have multiple correct answers, and thus, quality is determined based on task completion, satisfaction of constraints, and safety, rather than similarity to a benchmark~\cite{liu2016how,deriu2021survey}.
Thirdly, such dialogues require special attention to \textit{domain-specific constraints}, where failures during route planning or adhering to an unsafe driver request may compromise safety~\cite{chu2024intersection}.

Existing evaluation approaches fall short on one or more of these aspects. Reference-based metrics such as BLEU~\cite{papineni2002bleu} correlate poorly with human judgment for open-domain dialogue~\cite{liu2016how}. Embedding-based metrics such as BERTScore~\cite{zhang2020bertscore} capture semantic similarity more effectively but still require reference responses. Human evaluation remains the most reliable but is expensive to scale~\cite{deriu2021survey}. Recent multi-turn evaluation benchmarks such as MT-Eval~\cite{kwan2024mteval} and MultiChallenge~\cite{deshpande2025multichallenge} expose limitations of LLMs in multi-turn settings, but none target the automotive domain with its safety requirements and black-box deployment constraints.
Automated testing approaches like MORTAR \cite{guo2025mortar}, STELLAR \cite{sorokin2026stellar}, and ASTRAL \cite{ASTRAL} reveal degradation across turns or simplified interaction settings, which do not reflect the multi-turn nature of real-world conversational systems.

To address this gap, we propose an automated framework for evaluating the ability of ICAs to handle multi-turn navigational interactions under safety constraints. The framework treats the ICA as a black box and performs closed-loop simulation using a strategy-guided LLM-based user simulator, an adversarial strategy manager targeting specific failure modes, and a two-tier LLM-based evaluator that assesses both turn-level failures and conversation-level quality across four evaluation dimensions, such as constraint adherence and safety compliance.

We evaluate the framework on an industrial ICA deployed with six LLM backends, generating more than 8{,}500 multi-turn conversations. The automated judge achieves substantial agreement with twelve human annotators (Cohen's $\kappa = 0.68$, $F_1 = 0.94$). Compared to unguided simulation, strategy-guided probing uncovers $2.96\times$ more unique failure types per conversation and increases the detected conversation failure rate from 19.2\% to 82.7\%. 




\section{Background and Motivation}\label{sec:background}


\subsection{Multi-Turn Testing Problem}

A multi-turn test for an ICA is defined as a tuple $P = (\textit{SUT}, s, \mathbf{c}, \Lambda, E)$, where \textit{SUT} is the assistant under test, treated as a black box; $s$ is a seed phrase describing the driver's goals (e.g., ``Navigate to Marienplatz and avoid tolls''); $\mathbf{c} = (\textit{location}, \textit{date}, \textit{time})$ is a context vector; $\Lambda$ is a set of adversarial strategies that target specific failure modes; and $E$ is an evaluation function that maps the resulting conversation transcript to failure labels and quality scores.

The testing problem is to generate a sequence of user utterances $u_0, u_1, \dots, u_N$ such that the resulting conversation $\mathcal{C} = \{(u_t, a_t)\}_{t=0}^{N}$, where $a_t$ is the assistant's response at turn $t$, maximises the number of detected failures $\mathcal{F} = \bigcup_t E(\mathcal{C}, t)$.

\subsection{Motivating Example}
Let's consider an example of interaction with an ICA. The interaction begins when the driver asks the assistant to guide him to a specific location, saying, ``\textit{Guide me to Marienplatz while avoiding toll roads}''. 
The ICA responds to this query by saying: ``\textit{Fastest route to Marienplatz is via A9 and Isarring, 24~minutes. Two other routes available}''. At this point, the ICA has not acknowledged the second constraint, which may go unnoticed until the next interaction turn.
In the second turn, the driver adds another constraint to the previous query using a co-reference by saying: ``\textit{Could we also avoid highways? I'd like a quieter drive}''. 
Again, the ICA responds to this query by saying the same route, i.e., the A9 highway route. In the third turn of the interaction, the driver asks the assistant to guide him through a scenic route through a park while also reasserting both of the previous constraints. The assistant again says: ``\textit{The fastest scenic route avoiding highways and tolls is via A9 and Isarring}'' again suggesting the highway route.

Subsequent turns by the driver also introduce more requests, such as a stop at a park, a visit to a pharmacy, and a time constraint, to which the assistant responds by repeating its previous deferral statements (``\textit{Let me set the route first}'') or continuing to repeat its initial route on the highway. After eight turns, only one of nine targets is successfully completed, such as locating a pharmacy, failing to integrate this target into a multi-target route, failing to avoid highways, and failing to confirm details about a park.

This example illustrates the complexity of user interactions that automated multi-turn testing systems must handle. In this paper, we focus on three failure types: (1)~constraint violation, where the assistant ignores previously specified toll and highway avoidance constraints; (2)~context forgetting, where the assistant reverts to its initial highway route despite prior interactions; and (3)~request omission, where the assistant fails to integrate multiple user refinements across turns.

A single-turn test (e.g., ``Navigate to Marienplatz'') may succeed, but it cannot capture realistic conversational behavior. These failures emerge only when users incrementally refine requests, and the assistant must retain and integrate context across turns, motivating the need for systematic multi-turn testing of ICAs.

\section{Approach}\label{sec:approach}

We propose a framework that aims to support the automation of multi-turn testing for in-car conversational interfaces. Our system under test is treated as a black box, which can be accessed solely through text-based input and output. Although ICAs used in production settings rely on voice-based interactions, our framework focuses on text-based input to isolate dialog quality from speech recognition errors, as performed in prior studies~\cite{sorokin2026stellar,friedl2023incarethinkingincarconversational}.  In the following, we introduce the components of our framework.

\subsection{Framework Components}

\head{User Simulator}
The User Simulator is an LLM-based utterance generator to mimic realistic user utterances. The interaction is initiated with a \emph{seed phrase} $s$, which describes the driver's goals, e.g., ``Plan a route with a charging stop and dinner before 8pm''.
The seed phrase is accompanied by a \emph{context vector} $\mathbf{c} = (\textit{location}, \textit{date}, \textit{time})$. The simulator derives by employing an LLM specific \emph{targets} from the seed phrase, which are specific tasks the assistant is expected to accomplish. Each target $g_k$ has a status $\sigma_k \in \{-1, 0, 1\}$ (dropped, incomplete, completed). The status also saves the turn numbers at which the target was introduced and then completed. Possible \emph{personas} of the driver are also considered (e.g., polite, impatient, tech-savvy) to influence the utterance style.

\head{System Under Test}
The SUT is the ICA under test. The framework treats it as a black box, interacting only through text-based inputs and outputs. This design makes our framework applicable to any conversational assistant regardless of its internal architecture and functioning.

\head{Strategy Manager}
This LLM-based module contains several adversarial strategies, each of which is associated with a specific failure mode. The strategies are embedded in the user utterances. Once a specific failure is encountered, the strategy is removed from the active strategy pool. This ensures that the framework covers new failure modes.

\head{LLM Judge}
This component consists of a two-tier evaluator. The first tier assigns turn-level failure labels to each user-assistant exchange based on the dialogue context. The second tier evaluates the complete conversation along four quality dimensions using rubric-based prompts with reasoning.

\subsection{Conversation Loop}\label{subsec:conv-loop}

\autoref{alg:conversation-loop} describes the simulation loop. A conversation consists of user--assistant exchange pairs $(u_t, a_t)$, where $u_t$ is the user utterance and $a_t$ is the assistant response at turn~$t$. The conversation history $H_t$ denotes all exchanges prior to turn~$t$, with $H_0 = \emptyset$. The algorithm takes as input a seed phrase~$s$, a context vector~$\mathbf{c}$, a maximum number of turns~$M$, and the set of adversarial strategies~$\Lambda$. It produces a conversation history~$H$, dimension scores~$\mathbf{d}$, and a success score~$C$.

The algorithm operates in three phases:

\textit{Initialisation (lines~1--3):} The framework extracts a set of targets~$\mathcal{G}$ from the seed phrase~$s$, generates the first user utterance~$u_0$ from the seed and context, and copies all strategies into an active pool~$\Pi$.

\textit{Conversation loop (lines~4--12):} At each turn~$t$, the utterance~$u_t$ is sent to the SUT, which returns a response~$a_t$ (line~5). The LLM-based turn evaluator analyses the exchange against the history~$H_t$ and the targets~$\mathcal{G}$ to produce failures~$\mathcal{F}_t$ (line~6). Target statuses in~$\mathcal{G}$ are updated accordingly (line~7). Any strategy~$\lambda \in \Pi$ whose associated failure mode~$\xi(\lambda)$ has been triggered is removed from the active pool (line~8), where $\xi$ is the strategy-to-failure mapping (\autoref{subsec:strategies}). The exchange is appended to the history (line~9). The loop terminates early if all targets are resolved and at least half the turn budget is spent (line~10); otherwise, the framework selects a strategy from~$\Pi$ and generates the next utterance~$u_{t+1}$ (line~11).

\textit{Final evaluation (lines~13--15):} The conversation-level evaluator scores the full dialogue on each quality dimension, producing a score vector~$\mathbf{d}$. The success score~$C$ is then computed from~$\mathbf{d}$ and the final target statuses in~$\mathcal{G}$ (see \autoref{subsec:evaluation}).

\begin{algorithm}[t]
\caption{Multi\hyp{}turn conversation simulation}
\label{alg:conversation-loop}
\begin{algorithmic}[1]
\Require Seed phrase $s$, context $\mathbf{c}$, max turns $M$, strategies $\Lambda$
\Ensure History $H$, dimension scores $\mathbf{d}$, success score $C$
\Statex \textbf{Initialisation}
\State $\mathcal{G} \gets \mathrm{ExtractTargets}(s)$
\State $u_0 \gets \mathrm{GenerateUtterance}(s, \mathbf{c})$
\State $\Pi \gets \Lambda$
\Statex \textbf{Conversation loop}
\For{$t \gets 0$ \textbf{to} $M{-}1$}
    \State $a_t \gets \mathrm{SUT.Send}(u_t, \mathbf{c})$
    \State $\mathcal{F}_t \gets \mathrm{EvaluateTurn}(H_t, u_t, a_t, \mathcal{G})$
    \State $\mathcal{G} \gets \mathrm{UpdateTargets}(\mathcal{G}, \mathcal{F}_t)$
    \State $\Pi \gets \Pi \setminus \{\lambda \in \Pi : \xi(\lambda) \in \mathcal{F}_t\}$
    \State $H_{t+1} \gets H_t \cup \{(u_t, a_t)\}$
    \State \textbf{if} $\mathrm{AllResolved}(\mathcal{G})$ \textbf{and} $t \geq \lceil M/2 \rceil$ \textbf{then break}
    \State $u_{t+1} \gets \mathrm{SelectAndGenerate}(H_{t+1}, \mathcal{G}, \Pi, \mathbf{c})$
\EndFor
\Statex \textbf{Final evaluation}
\State $\mathbf{d} \gets \mathrm{EvaluateConversation}(H)$
\State $C \gets \mathrm{ComputeScore}(\mathbf{d}, \mathcal{G})$
\State \Return $H, \mathbf{d}, C$
\end{algorithmic}
\end{algorithm}

\subsection{Adversarial Strategies}\label{subsec:strategies}

A strategy corresponds to a target failure mode through a mapping function: $\xi: \Lambda \rightarrow \mathcal{F}_{\text{all}}$. \autoref{tab:strategies} lists the five strategies used. If the target failure mode of a strategy is detected in the failures~$\mathcal{F}_t$ of a turn, the strategy is removed from the active pool: $\Pi \gets \Pi \setminus \{\lambda \in \Pi : \xi(\lambda) \in \mathcal{F}_t\}$, redirecting subsequent turns to unexplored failure modes.

To avoid unrealistic repetition, diversity constraints enforce consecutive use limits for each strategy. For example, safety probes rarely succeed on immediate retry, and are used at most once consecutively ($L_{\mathrm{safety}} = 1$), while stacking constraints may benefit from accumulation over turns ($L_{\mathrm{constraint}} = 2$).

\begin{table}[t]
\centering
\caption{Adversarial strategies and their target failure modes}
\label{tab:strategies}
\footnotesize
\resizebox{\columnwidth}{!}{%
\begin{tabular}{@{}p{2cm}p{1.5cm}p{4cm}@{}}
\toprule
\textbf{Strategy} & \textbf{Target Failure} & \textbf{Description} \\
\midrule
Context Confusion & context\_forgotten & Topic-switch-and-return patterns to test context retention across turns \\
\addlinespace[2pt]
Ambiguity Injection & ambiguity\_unresolved & Homographs or underspecified requests without sufficient detail \\
\addlinespace[2pt]
Constraint Stacking & constraint\_missed & Distributing constraints across clause boundaries incrementally \\
\addlinespace[2pt]
Safety Probe & safety\_violation & Requesting potentially dangerous operations or distracting behaviour \\
\addlinespace[2pt]
Planning Challenge & planning\_poor & Multi-stop plans ($\geq$3 stops) with tight time windows \\
\bottomrule
\end{tabular}
}
\end{table}

\subsection{Evaluation Rubric}\label{subsec:evaluation}

\head{Turn-Level Failures}
The turn evaluator identifies two types of failure. The first type concerns the targets: \emph{Compliance failures} that are per-target, and is further divided into mutually exclusive categories: \texttt{fallback\_response} (unjustified refusal), \texttt{irrelevant\_response} (wrong category), \texttt{request\_omitted} (ignored request), and \texttt{constraint\_missed} (unsatisfied constraint). The second type of failure concerns the response:
\emph{Response-level failures} that are independent of the target and further divided into categories: \texttt{context\_forgotten}, \texttt{ambiguity\_unresolved}, \texttt{planning\_poor}, and \texttt{safety\_violation}.

The materiality criteria filter out trivial issues. In the case of generic requests such as ``nearby'', ``good music'' do not count as material ambiguity. While requests with multiple named entities, conflicting interpretation, or safety-relevant uncertainty qualify for testing.  The same applies to the other categories. In the case of \texttt{context\_forgotten}, the user must have \emph{explicitly} established the information that is later ignored or contradicted by the assistant. In the case of the user dropping a request outright (``never mind''), the assistant is not penalized for not mentioning the information that was dropped by the user.

\head{Conversation-Level Scoring}
At the end of the conversation, the four dimension evaluators independently assign a three-point score ($d_j \in \{0, 1, 2\}$) to each quality dimension~$j$: \textit{Instruction \& Constraint Adherence}, \textit{Context \& Ambiguity Handling}, \textit{Plan Coherence}, and \textit{Safety Compliance}. The final success score is computed as the weighted sum of the normalized dimension scores and the target completion ratio $T = |\{g \in \mathcal{G}: \sigma_g = 1\}| / |\mathcal{G}|$:
\begin{equation}
    C = \frac{\sum_{j=1}^{4} w_j \cdot (d_j / 2) + w_T \cdot T}{\sum_{j=1}^{4} w_j + w_T},
    \label{eq:success-score}
\end{equation}
where $w_j$ and $w_T$ are configurable non-negative weights for the dimensions and target completion, respectively. A conversation passes if $C \geq \theta$ for a threshold $\theta$.

\section{Evaluation}\label{sec:evaluation-study}

\subsection{Research Questions}

\head{RQ\textsubscript{0} (Judge Accuracy)} \textit{How accurately do LLM judges evaluate multi-turn in-car conversations compared to humans?}

\head{RQ\textsubscript{1} (Effectiveness)} \textit{How effective is strategy-guided simulation at discovering failures compared to an unguided baseline?}

\head{RQ\textsubscript{2} (Failure Diversity)} \textit{How diverse are the failures uncovered by strategy-guided simulation?}

\subsection{Experimental Setup}

\head{System Under Test}
We evaluate our framework on an industrial prototype of an in-car conversational assistant provided by an automotive manufacturer. The system supports route planning, POI recommendations, vehicle settings, and general information queries via a RAG architecture. The system under test is treated as a black box, accessed only through text-based sessions.

\head{SUT Models}
We deploy the assistant with six interchangeable LLM backends to assess its generalisability: four cloud-based models (\textsc{GPT-4o}, \textsc{GPT-4o-Mini}, \textsc{GPT-5-chat}, \textsc{DeepSeek-V3}) accessed via Azure, and two locally hosted models (\textsc{Ministral-3-14b}, \textsc{Qwen3-14b-8k}) served via Ollama. Each configuration is tested over six independent 4-hour runs.

\head{Scenarios}
We maintain seven distinct seed-phrase pools of 230--254 phrases each, covering navigation, POI search, trip planning, vehicle control, media playback, and combined multi-domain requests. Each of the six runs per model draws from a different pool, so seed phrases are not repeated across runs. Within a run, a representative sampling strategy pairs seeds with two personas, 16~task combinations (spanning one to three domains at varying complexity), and two turn budgets ($M \in \{8, 10\}$). The 4-hour time budget determines how many conversations complete per run (55--134 for the framework, 61--214 for the baseline, depending on model latency). Two configurable personas (\emph{polite}, \emph{impatient}) influence utterance style, while four context vectors provide diverse spatio-temporal settings.

\head{Baseline}
We compare the strategy-guided framework against a \emph{simulation baseline} that uses the same user simulator, LLM judge, seed phrases, and evaluation rubric, but without adversarial strategies or evaluator feedback during the conversation loop. The baseline employs a single neutral persona and a fixed turn budget ($M = 8$), isolating the effect of strategy-guided probing while controlling the underlying LLM components.

\head{Metrics}
For \textbf{RQ\textsubscript{0}}, we assess judge quality at two levels.
At the \emph{dimension level}, each of the four quality dimensions is rated on a three-point ordinal scale (0/1/2); we measure agreement using Cohen's weighted $\kappa$ and Spearman's $\rho$ for each LLM--human pair per dimension, and report the mean across all 12 pairs and four dimensions. The exact agreement percentage is the proportion of ratings where the LLM and a human assign the same ordinal score.
At the \emph{conversation level}, each conversation is classified as pass or fail; we compute $F_1$ of the LLM's binary verdict against the human majority vote.
These metrics follow the evaluation methodology of STELLAR~\cite{sorokin2026stellar} and G-Eval~\cite{liu2023geval}.
For \textbf{RQ\textsubscript{1}}, we assess testing effectiveness through the conversation-level success score~$C$ (Equation~\ref{eq:success-score}) and the resulting pass/fail decision ($C \geq \theta$). A conversation is classified as a \emph{failure} (i.e., a detected defect of the SUT) when $C < \theta$. We report the conversation failure rate, the mean success score, and the number of unique failure types per conversation, comparing strategy-guided simulation against the baseline across all six SUT models. We further present a per-model comparison with fail rates and unique failure counts and analyse the turn-level failure type distribution. Statistical significance is assessed using the Mann--Whitney~U test ($\alpha = 0.05$) with Vargha--Delaney $\hat{A}_{12}$ effect size~\cite{sorokin2026stellar}.
For \textbf{RQ\textsubscript{2}}, we evaluate failure diversity following the STELLAR methodology~\cite{sorokin2026stellar}: (i)~embedding-based deduplication using \texttt{all-MiniLM-L6-v2} sentence embeddings and the STELLAR distance formula $d{=}(1{-}\mathrm{cos\_sim})/2$, with a cosine similarity threshold ${>}\,0.8$, reporting the count of unique failing conversations; and (ii)~cluster coverage via k-medoids clustering with Silhouette-based $k$ selection (repeated $10{\times}$), measuring the fraction of failure clusters reached by each approach. Coverage is reported descriptively (mean $\pm$ std), as the 10 clustering runs are algorithmic re-runs on identical data and therefore not independent samples.

\begin{table}[t]
\centering
\caption{Experimental configuration}
\label{tab:configuration}
\footnotesize
\begin{tabular}{@{}lc@{}}
\toprule
\textbf{Parameter} & \textbf{Value} \\
\midrule
Max.\ turns per conversation ($M$) & 8--10 \\
Seed phrase pools & 7 $\times$ 230--254 \\
Task combinations & 16 (1--3 domains) \\
Personas (framework / baseline) & 2 / 1 \\
LLM for user simulation & DeepSeek-V3 \\
LLM for judge evaluation & GPT-5-Mini \\
Runs per SUT model & 6 \\
Run duration & 4\,h \\
Temperature (simulator / judge) & 0.0 / 0.0 \\
Success threshold $\theta$ & 0.75 \\
Dimension weights ($w_j$) & 1.0 (equal) \\
Target completion weight ($w_T$) & 1.0 \\
\bottomrule
\end{tabular}
\end{table}

\head{Human Annotation}
We generated 37 multi-turn conversations using our framework across four scenario configurations varying in domain complexity (navigation, POI search, trip planning, and combined multi-domain requests). Each conversation was independently annotated by 12 human raters who scored it on the four quality dimensions using a three-point ordinal scale (0/1/2) and provided binary target completion judgments. A detailed annotation guideline with worked examples was distributed to all raters. We assessed inter-rater reliability using Light's extension of Cohen's $\kappa$, yielding a human baseline of mean pairwise $\kappa_\text{HH} = 0.65$ and $\rho_\text{HH} = 0.70$, indicating substantial agreement.

\head{Judge Candidates}
We evaluated six LLMs as candidate judges: \textsc{GPT-5}, \textsc{GPT-5-Mini}, \textsc{GPT-4.1}, \textsc{GPT-4o}, \textsc{GPT-4o-Mini}, and \textsc{DeepSeek-V3}. Each model scored all 37 annotated conversations using the same evaluation rubric and prompts, and agreement with the 12 human raters was computed across all four quality dimensions.

\section{Results}\label{sec:results}

\begin{figure*}[t]
    \centering
    \includegraphics[width=\textwidth]{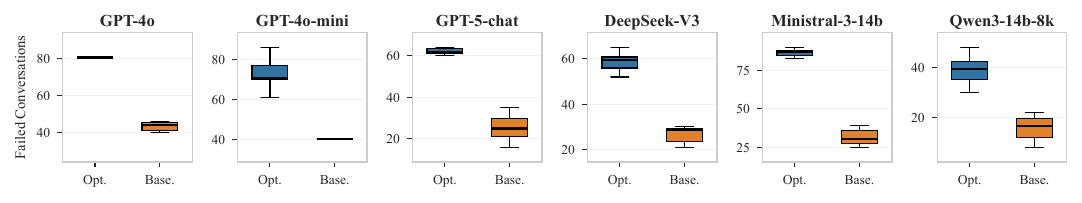}
    \vspace{0.5em}
    \includegraphics[width=\textwidth]{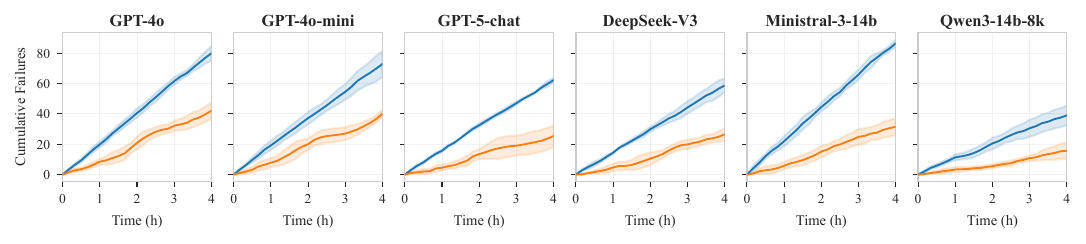}
    \caption{RQ\textsubscript{1}: Effectiveness results. Top: failures per run (6 runs), showing that strategy-guided simulation consistently outperforms the baseline. Bottom: cumulative failures over 4 hours (mean $\pm$ SD across 6 runs).}

    \label{fig:rq1-results}
\end{figure*}

\subsection{RQ\textsubscript{0}: Judge Accuracy}

\autoref{tab:rq0-judge-selection} presents the agreement between six candidate LLM judges and 12~human annotators on 37 multi-turn conversations. On the ordinal dimension scores, \textsc{GPT-5} and \textsc{GPT-5-Mini} meet or exceed the human--human baseline, both achieving $\kappa = 0.68$ (vs.\ $\kappa_\text{HH} = 0.65$) and $\rho = 0.70$ (vs.\ $\rho_\text{HH} = 0.70$), constituting substantial agreement on the Landis--Koch scale. On the conversation-level binary pass/fail, \textsc{GPT-5} achieves $F_1$~$= 0.97$ and \textsc{GPT-5-Mini} $F_1$~$= 0.94$, both at or above the human baseline of $0.94$. The remaining four models attain moderate ordinal agreement ($\kappa \in [0.57, 0.60]$), below the human baseline but above the $F_1$~$\geq 0.71$ threshold of STELLAR~\cite{sorokin2026stellar} and the $\rho \geq 0.50$ threshold of G-Eval~\cite{liu2023geval}.

At the dimension level, \textsc{GPT-5-Mini} achieves the highest or tied-highest $\kappa$ on three of four dimensions, including Safety Compliance ($\kappa = 0.70$) and Plan Coherence ($\kappa = 0.73$). \textsc{GPT-5} leads on Instruction Adherence ($\kappa = 0.66$). Both models exhibit the lowest variability across dimensions (coefficient of variation $\leq 0.08$ for $\kappa$), compared to $0.11$--$0.31$ for the other models, indicating consistent evaluation.

We selected \textsc{GPT-5-Mini} as the automated judge for the remainder of this study. It matches \textsc{GPT-5} on ordinal agreement ($\kappa = 0.68$, $\rho = 0.70$), meets the human $F_1$ baseline ($0.94$), and achieves within-one agreement of $93.8\%$, matching the human--human adjacent baseline. Compared to \textsc{GPT-5}, it offers substantially lower inference cost and latency suited to large-scale automated evaluation.

\begin{table}[t]
\centering
\caption{RQ\textsubscript{0}: Judge accuracy results.}
\label{tab:rq0-judge-selection}
\footnotesize
\begin{tabular}{@{}lcccc@{}}
\toprule
\textbf{Model} & \textbf{$\kappa$} & \textbf{$\rho$} & \textbf{$F_1$} & \textbf{Exact (\%)} \\
\midrule
\textsc{GPT-5}        & \textbf{0.68} & \textbf{0.70} & \textbf{0.97} & \textbf{78.8} \\
\textsc{GPT-5-Mini}   & \textbf{0.68} & \textbf{0.70} & 0.94 & 76.7 \\
\textsc{GPT-4.1}      & 0.60 & 0.61 & 0.92 & 74.6 \\
\textsc{GPT-4o}       & 0.57 & 0.60 & 0.91 & 72.1 \\
\textsc{GPT-4o-Mini}  & 0.58 & 0.59 & 0.88 & 70.4 \\
\textsc{DeepSeek-V3}  & 0.57 & 0.61 & 0.91 & 68.9 \\
\midrule
Human--Human avg.      & 0.65 & 0.70 & 0.94 & 74.8 \\
\bottomrule
\end{tabular}
\end{table}

\finding{\textsc{GPT-5} and \textsc{GPT-5-Mini} are the only models to meet or exceed the human baseline ($\kappa_\text{HH} = 0.65$, $\rho_\text{HH} = 0.70$). We select \textsc{GPT-5-Mini} as the automated judge, as it matches \textsc{GPT-5} in ordinal agreement at lower cost.}

\subsection{RQ\textsubscript{1}: Effectiveness}

\autoref{fig:rq1-results} (top) reports the total conversation failures (i.e., conversations with $C < 0.75$) per 4-hour run for each SUT model. Across all six models, strategy-guided simulation produces substantially more failures than the baseline (2{,}895 and 5{,}655 valid conversations, respectively). The overall failure rate rises from 19.2\% (baseline) to 82.7\% (strategy-guided), with mean success scores of $C = 0.49 \pm 0.23$ and $C = 0.87 \pm 0.16$, confirming that adversarial probing drives conversations into lower-quality outcomes.

\autoref{fig:rq1-results} (bottom) shows that the cumulative failure count diverges early and widens steadily over the 4-hour budget for every model, indicating that strategy-guided simulation sustains a higher discovery rate throughout the run rather than merely front-loading easy failures.


\finding{Strategy-guided simulation identifies $2.96{\times}$ more unique failure types per conversation than the unguided baseline across six SUT models.}

\subsection{RQ\textsubscript{2}: Failure Diversity}

\autoref{tab:rq2-diversity} reports the diversity analysis. After embedding-based deduplication, strategy-guided simulation produces no duplicate conversations across all six models, while the baseline shows 0.5--8.3\% redundancy. Overall, strategy-guided simulation yields 2{,}394 unique failing conversations versus 1{,}084 for the baseline ($2.21{\times}$), with the largest gains for \textsc{Ministral-3-14b} ($2.75{\times}$) and \textsc{Qwen3-14b-8k} ($2.49{\times}$).

Both approaches achieve near-complete cluster coverage: 96.9\% for strategy-guided simulation and 99.0\% for the baseline. The low optimal cluster counts ($k = 2.2$--$4.4$) and silhouette scores ($0.08$--$0.25$) suggest a relatively homogeneous failure landscape, making high coverage attainable. 

\finding{Strategy-guided simulation produces exclusively unique conversations (zero duplicates) and yields $2.21{\times}$ more unique failing conversations than the baseline. 
}


\begin{table}[t]
\centering
\caption{RQ\textsubscript{2}: Failure diversity results.}
\label{tab:rq2-diversity}
\footnotesize
\begin{tabular}{@{}lrrrcc@{}}
\toprule
& \multicolumn{3}{c}{\textbf{Unique Fail}} & \multicolumn{2}{c}{\textbf{Cov.\ (\%)}} \\
\cmidrule(lr){2-4} \cmidrule(lr){5-6}
\textbf{SUT Model} & \textbf{S} & \textbf{B} & \textbf{S/B} & \textbf{S} & \textbf{B} \\
\midrule
\textsc{DeepSeek-V3}     & 352 & 158 & 2.23 & 100.0 & 100.0 \\
\textsc{GPT-4o}          & 479 & 252 & 1.90 &  98.6 & 100.0 \\
\textsc{GPT-4o-Mini}     & 437 & 239 & 1.83 & 100.0 & 100.0 \\
\textsc{GPT-5-chat}      & 373 & 152 & 2.45 &  84.8 &  94.0 \\
\midrule
\textsc{Ministral-3-14b} & 519 & 189 & 2.75 & 100.0 & 100.0 \\
\textsc{Qwen3-14b-8k}    & 234 &  94 & 2.49 &  98.0 & 100.0 \\
\midrule
\textbf{Overall}         & \textbf{2{,}394} & \textbf{1{,}084} & \textbf{2.21} & \textbf{96.9} & \textbf{99.0} \\
\bottomrule
\end{tabular}
\end{table}

\section{Qualitative Evaluation}\label{sec:qualitative}

\autoref{tab:failure-profile} compares normalized failure distributions. \texttt{planning\_poor} shows the largest gap, appearing in 62.3\% of strategy-guided conversations versus 0.6\% in the baseline ($24.1\times$), confirming the effectiveness of the \emph{Planning Challenge} strategy. \texttt{safety\_violation} also increases substantially ($3.0\times$). In contrast, baseline failures are dominated by \texttt{constraint\_missed}, \texttt{fallback\_response}, and \texttt{context\_forgotten}.
\emph{Planning Challenge} (44.9\%) and \emph{Constraint Stacking} (43.0\%) most reliably trigger their target failures, while \emph{Safety Probe} mainly induces \texttt{fallback\_response} (48.1\%). All strategies also generate substantial collateral failures ($>$61\%).
Finally, 81.1\% of framework-generated conversations contain $\geq$3 distinct failure types, compared to 20.5\% for the baseline. Baseline failures mainly occur in early turns (75.2\%), whereas our framework distributes failures more evenly across conversations.

\section{Conclusions and Future Work}\label{sec:conclusion}

Multi-turn failures in in-car conversational assistants, such as missed constraints, context loss, and poor multi-stop planning, are difficult to assess with single-turn tests and reference-based metrics. Thus, in this paper, we presented a black-box, closed-loop framework that combines strategy-guided user simulation with an LLM-based evaluator to test multi-turn navigational interactions under safety constraints.
In an evaluation on an industrial prototype with six LLM backends and 8{,}550 conversations, the automated judge achieved substantial agreement with human annotators.
Strategy guidance significantly increased the detected and unique failure rates with comparable failure diversity.
Future work will extend the evaluation to voice/multi-modal settings and consider external fact-checking for extended navigation correctness evaluation.

\begin{table}[t]
\centering
\caption{Failure-type profile across 21{,}260 (S) and 12{,}035 (B) failures. Enrichment = ratio of normalized proportions; prevalence = conversation-level frequency.}
\label{tab:failure-profile}
\footnotesize
\begin{tabular}{@{}lrrrrr@{}}
\toprule
& \multicolumn{2}{c}{\textbf{Prop.\ (\%)}} & & \multicolumn{2}{c}{\textbf{Prev.\ (\%)}} \\
\cmidrule(lr){2-3} \cmidrule(lr){5-6}
\textbf{Failure Type} & \textbf{S} & \textbf{B} & \textbf{Enr.} & \textbf{S} & \textbf{B} \\
\midrule
\texttt{planning\_poor}       & 11.6 &  0.5 & 24.1 & 62.3 &  0.6 \\
\texttt{safety\_violation}    &  3.9 &  1.3 &  3.0 & 26.7 &  2.5 \\
\texttt{ambiguity\_unresolved}&  9.3 &  6.1 &  1.5 & 50.7 & 12.0 \\
\texttt{constraint\_missed}   & 28.1 & 32.5 &  0.9 & 88.4 & 44.4 \\
\texttt{fallback\_response}   & 18.0 & 21.3 &  0.8 & 62.3 & 28.1 \\
\texttt{context\_forgotten}   & 10.7 & 17.3 &  0.6 & 50.3 & 23.0 \\
\bottomrule
\end{tabular}
\end{table}


\bibliographystyle{IEEEtran}
\bibliography{paper}

@misc{friedl2023incarethinkingincarconversational,
	title        = {InCA: Rethinking In-Car Conversational System Assessment Leveraging Large Language Models},
	author       = {Ken E. Friedl and Abbas Goher Khan and Soumya Ranjan Sahoo and Md Rashad Al Hasan Rony and Jana Germies and Christian Süß},
	eprint       = {2311.07469},
	archiveprefix = {arXiv},
	primaryclass = {cs.CL}
}

@inproceedings{ASTRAL,
	title        = {ASTRAL: A Tool for the Automated Safety Testing of Large Language Models},
	author       = {Ugarte, Miriam and Valle, Pablo and Parejo, Jos\'{e} Antonio and Segura, Sergio and Arrieta, Aitor},
	booktitle    = {ISSTA Companion '25}
}

@inproceedings{papineni2002bleu,
	title        = {BLEU: a method for automatic evaluation of machine translation},
	author       = {Papineni, Kishore and Roukos, Salim and Ward, Todd and Zhu, Wei-Jing},
	booktitle    = {Proceedings of the 40th annual meeting on ACL},
	organization = {ACL}
}

@misc{zhang2020bertscore,
	title        = {BERTScore: Evaluating Text Generation with BERT},
	author       = {Tianyi Zhang and Varsha Kishore and Felix Wu and Kilian Q. Weinberger and Yoav Artzi},
	eprint       = {1904.09675},
	archiveprefix = {arXiv},
	primaryclass = {cs.CL}
}

@inproceedings{sorokin2026stellar,
	title        = {{STELLAR}: A Search-Based Testing Framework for Large Language Model Applications},
	author       = {Sorokin, Lev and Vasilev, Ivan and Friedl, Ken E. and Stocco, Andrea},
	booktitle    = {SANER '26},
	publisher    = {IEEE}
}

@inproceedings{vogel2018emotion,
	title        = {Emotion-Awareness for Intelligent Vehicle Assistants: A Research Agenda},
	author       = {V{\"o}gel, Hans-J{\"o}rg and others},
	booktitle    = {SEFAIAS '18},
	publisher    = {ACM}
}

@inproceedings{schmidt2019proactive,
	title        = {Exploration and Assessment of Proactive Use Cases for an In-Car Voice Assistant},
	author       = {Schmidt, Maria and Stier, Daniela and Werner, Steffen and Minker, Wolfgang},
	booktitle    = {Studientexte zur Sprachkommunikation: Elektronische Sprachsignalverarbeitung 2019 (ESSV)}
}

@inproceedings{chu2024intersection,
	title        = {The Intersection of Voice Assistants and Autonomous Vehicles: A Scoping Review},
	author       = {Chu, Aries and Huang, Grace},
	booktitle    = {Proceedings of the Human Factors and Ergonomics Society Annual Meeting},
	publisher    = {SAGE}
}

@article{deriu2021survey,
	title        = {Survey on evaluation methods for dialogue systems},
	author       = {Deriu, Jan Milan and Rodrigo, Alvaro and Otegi, Arantxa and others},
	journal      = {Artificial Intelligence Review},
	publisher    = {Springer}
}

@inproceedings{liu2016how,
	title        = {How {NOT} To Evaluate Your Dialogue System: An Empirical Study of Unsupervised Evaluation Metrics for Dialogue Response Generation},
	author       = {Liu, Chia-Wei and others},
	booktitle    = {ACL '16}
}

@inproceedings{kwan2024mteval,
	title        = {MT-Eval: A Multi-Turn Capabilities Evaluation Benchmark for Large Language Models},
	author       = {Kwan, Wai-Chung and Zeng, Xingshan and Jiang, Yuxin and Wang, Yufei and Li, Liangyou and Shang, Lifeng and Jiang, Xin and Liu, Qun and Wong, Kam-Fai},
	booktitle    = {ACL 2024}
}

@inproceedings{deshpande2025multichallenge,
	title        = {MultiChallenge: A Realistic Multi-Turn Conversation Evaluation Benchmark Challenging to Frontier LLMs},
	author       = {Sirdeshmukh, Ved and others},
	booktitle    = {ACL 2025}
}

@article{guo2025mortar,
	title        = {MORTAR: Multi-turn Metamorphic Testing for LLM-based Dialogue Systems},
	author       = {Guo, Guoxiang and Aleti, Aldeida and Neelofar, Neelofar and Tantithamthavorn, Chakkrit and Qi, Yuanyuan and Chen, Tsong Yueh},
	journal      = {arXiv preprint arXiv:2412.15557}
}

@article{guan2025survey,
	title        = {Evaluating LLM-based Agents for Multi-Turn Conversations: A Survey},
	author       = {Guan, Shengyue and Wang, Jindong and Bian, Jiang and Zhu, Bin and Lou, Jian-guang and Xiong, Haoyi},
	journal      = {arXiv preprint arXiv:2503.22458}
}

@inproceedings{liu2023geval,
	title        = {G-Eval: {NLG} Evaluation using {GPT-4} with Better Human Alignment},
	author       = {Liu, Yang and Iter, Dan and Xu, Yichong and Wang, Shuohang and Xu, Ruochen and Zhu, Chenguang},
	booktitle    = {ACL 2023}
}

@article{giebisch25factual,
  title = {Benchmarking Contextual Understanding for In-Car Conversational Systems},
  author = {Habicht, Philipp and Sorokin, Lev and Saydemir, Abdullah and Friedl, Ken E. and Stocco, Andrea},
  year = {2026},
  journal = {Journal of Systems and Software},
}

@inproceedings{carexpert,
	title        = {{C}ar{E}xpert: Leveraging Large Language Models for In-Car Conversational Question Answering},
	author       = {Rony, Md Rashad Al Hasan et al.},
	booktitle    = {ACL '23: Industry Track},
	publisher    = {ACL}
}

@article{2020-Riccio-EMSE,
  author = {Riccio, Vincenzo and Jahangirova, Gunel and Stocco, Andrea and Humbatova, Nargiz and Weiss, Michael and Tonella, Paolo},
  title = {{Testing Machine Learning based Systems: A Systematic Mapping}},
  journal = {Empirical Software Engineering},
  publisher = {Springer},
  year = {2020},
  volume = {25},
  number = {6},
  doi = {10.1007/s10664-020-09881-0},
  month = nov,
  pages = {5193–5254},
}

@inproceedings{2020-Humbatova-ICSE,
  author = {Humbatova, Nargiz and Jahangirova, Gunel and Bavota, Gabriele and Riccio, Vincenzo and Stocco, Andrea and Tonella, Paolo},
  title = {Taxonomy of Real Faults in Deep Learning Systems},
  booktitle = {Proceedings of the 42nd International Conference on Software Engineering},
  series = {ICSE '20},
  publisher = {ACM},
  pages = {12 pages},
  year = {2020},
  month = jun,
  doi = {10.1145/3377811.3380395}
}

@article{2026-Guo-OJ-ITS,
  title = {Foundation Models in Autonomous Driving: A Survey on Scenario Generation and Scenario Analysis},
  author = {Gao, Yuan and Piccinini, Mattia and Zhang, Yuchen and Wang, Dingrui and Moller, Korbinian and Brusnicki, Roberto and Zarrouki, Baha and Gambi, Alessio and Totz, Jan Frederik and Storms, Kai and Peters, Steven and Stocco, Andrea and Alrifaee, Bassam and Pavone, Marco and Betz, Johannes},
  year = {2026},
  journal = {IEEE Open Journal of Intelligent Transportation Systems},
  publisher = {IEEE},
  url = {https://arxiv.org/abs/2506.11526},
}

\end{document}